\documentclass[
]{ceurart}

\usepackage{listings}
\usepackage{epigraph}
\usepackage{amsmath}
\usepackage{amssymb}
\usepackage{amsthm}
\theoremstyle{definition}

\theoremstyle{plain}

\theoremstyle{remark}

\theoremstyle{assumption}

\usepackage{tikz}
\usetikzlibrary{arrows.meta,positioning,calc,fit,backgrounds,decorations.pathreplacing}
\newtheorem{constraint}{Constraint}[section]
\usepackage[most]{tcolorbox}
\usepackage{xcolor}

\usepackage{hyperref} 

\definecolor{NorthwesternPurple}{RGB}{92, 64, 140}

\definecolor{ModernBlue}{RGB}{38, 84, 124}

\definecolor{ModernRed}{RGB}{156, 47, 57}

\definecolor{ModernPurple}{RGB}{92, 64, 140}

\colorlet{AxiomRed}{ModernRed}
\colorlet{MetaBlue}{ModernBlue}

\newcommand{\coloredheadnum}[3]{%
  #2{%
    \texorpdfstring{%
      {\begingroup
       \color{#1}\bfseries #3%
       \endgroup}%
    }{#3}%
  }%
}

\newcommand{\axiomtag}[1]{%
  \texorpdfstring{%
    \begingroup
    \tcbox[
      on line,
      colback=AxiomRed,
      colframe=AxiomRed,
      arc=2pt,              
      boxrule=0pt,          
      left=2pt,
      right=2pt,
      top=1pt,
      bottom=1pt,
    ]{\color{white}\bfseries\scshape #1}%
    \endgroup
  }{#1}%
}

\newcommand{\metatag}[1]{%
  \texorpdfstring{%
    \begingroup
    \tcbox[
      on line,
      colback=MetaBlue,
      colframe=MetaBlue,
      arc=2pt,              
      boxrule=0pt,          
      left=2pt,
      right=2pt,
      top=1pt,
      bottom=1pt,
    ]{\color{white}\bfseries\scshape #1}%
    \endgroup
  }{#1}%
}

\newcommand{\axiomsubsection}[1]{%
  \coloredheadnum{AxiomRed}{\subsection}{#1}%
}

\newcommand{\metasubsection}[1]{%
  \coloredheadnum{MetaBlue}{\subsection}{#1}%
}

\begin{document}

\copyrightyear{2026}
\copyrightclause{Copyright for this paper by its authors.
  Use permitted under Creative Commons License Attribution 4.0
  International (CC BY 4.0).}

\conference{AAAI-26 Workshop on Machine Ethics: from formal methods to emergent machine ethics, January 27, 2026, Singapore}

\title{Constructive Alignment}
\title[mode=sub]{Governing Preference Dynamics in Human–AI Interaction}

\author[1]{Max Kanwal}[%
orcid=0009-0008-7021-3227,
email=kanwal@stanford.edu,
url=https://linkedin.com/in/mkanwal/,
]
\fnmark[1]
\cormark[1]
\address[1]{Stanford University}

\author[2]{Caryn Tran}[%
orcid=0000-0002-4645-6607,
email=caryn@u.northwestern.edu,
url=https://linkedin.com/in/caryntran/,
]
\fnmark[1]
\address[2]{Northwestern University}

\fntext[1]{These authors contributed equally and are listed in alphabetical order.}
\cortext[1]{Corresponding author.}

\begin{abstract}
    Most approaches to AI alignment treat human preferences as fixed targets to be inferred and optimized. This assumption conflicts with extensive empirical evidence showing that preferences are layered, dynamic, and constructed through interaction—particularly with adaptive technologies. As AI systems become more persistent, personalized, and socially embedded, they increasingly participate in shaping what people attend to, value, and endorse over time. We introduce \textit{Constructive Alignment}, a paradigm that reframes alignment as a control problem over evolving human preference trajectories rather than static preference satisfaction. Drawing on behavioral economics, psychology, and constructivist social theory, we model preferences as layered state variables that evolve under interaction with AI systems. We formalize this view using a control-theoretic framework in which system actions and interaction design jointly influence both world states and human evaluative states. We argue that alignment is not primarily about controlling AI behavior, but about regulating how AI systems influence the evolution of human preferences—ensuring that value trajectories remain coherent, reflectively endorsed, epistemically grounded, bounded against manipulation, and empowering under uncertainty. Alignment thus becomes a problem of governing long-term value formation rather than simply satisfying static preferences.
\end{abstract}


\maketitle

\epigraph{\emph{“We shape our tools and thereafter our tools shape us.”}}{--- Marshall McLuhan}

\begin{figure}[h!]
    \centering
    \includegraphics[width=1.\linewidth]{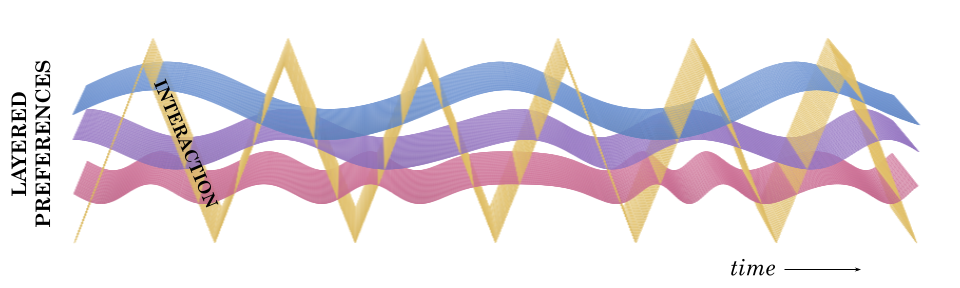}
    \caption{Constructive Alignment argues for consideration of the dynamic and constructed nature of preferences. Preferences are depicted as layered, continuous processes that evolve over time (blue, purple, pink), rather than a single static objective to satisfy. Interaction (yellow), including with AI, intermittently intervenes, reshaping these layers, illustrating how preferences are constructed and updated through experience.}
    \label{fig:preferences}
\end{figure}

\section{Introduction}

Over the course of a single weekend, YouTube recommendations can turn a casual listener into an obsessed fan. Users can develop emotional dependence on LLM-based AI therapists and AI romantic partners as primary sources of support \citep{heritage_i_2025, meadi_exploring_2025}. Political campaigns have shown how data-driven advertising can shape voter attitudes at scale \citep{greenfield_cambridge_2018}. Meanwhile, the internet may be eroding our capacity for sustained reading and focused attention \citep{firth_online_2019}. Across these cases, a common pattern emerges: AI systems do not merely respond to human preferences, but participate in shaping what people attend to, value, and come to want. As these systems become more capable, personalized, and persistent, their role in shaping human preferences and values becomes unavoidable.

Despite this, much of the AI alignment literature continues to frame alignment as a problem of matching system behavior to human preferences as they are. Preferences are treated as targets to be inferred, aggregated, or optimized against, implicitly assumed to be stable, well-defined, and external to the system. This framing leaves a critical dimension underspecified: how preferences themselves change over time, and how AI systems participate in that change.

Recent work has begun to recognize this gap. Scholars across AI ethics and alignment argue that preferences are socially embedded, context-dependent, and shaped by interaction \citep{franklin_recognising_2022, zhi-xuan_beyond_2024, shen_position_2025}. However, most existing approaches still focus on which values should count as alignment targets, rather than on the processes through which values are formed, revised, and stabilized under sustained interaction with AI systems.

In this paper, we argue that this omission is not merely a descriptive oversight but a structural limitation of prevailing alignment paradigms. Human preferences are layered across time horizons, dynamic across contexts and life stages, and constructed through interaction with environments, institutions, and technologies. Because AI systems inevitably influence these processes, alignment cannot be understood solely as an optimization problem over fixed objectives. It must also address how systems shape the conditions under which human preferences emerge and evolve.

We introduce Constructive Alignment as a paradigm that takes this challenge seriously. The term constructive draws on constructivism as an intellectual paradigm across psychology, learning sciences, decision research, and the social sciences. Constructivist accounts reject the view that knowledge—including preferences and goals—is a fixed internal representation that is passively acquired or merely revealed through observation or choice, emphasizing instead that they are actively formed through interaction. This paradigm appears in developmental and sociocultural psychology (e.g., \citet{piaget_origins_1952}, \citet{vygotskij_mind_1981}), in cognitive science and human--computer interaction (e.g., \citet{suchman_plans_1987}, \citet{hutchins_cognition_2006}), in the learning sciences (e.g., \citet{sawyer_cambridge_2009}), and in decision research and behavioral economics, where preferences are understood as constructed rather than revealed (e.g., \citet{slovic_construction_1995}, \citet{lichtenstein_construction_2006}, \citet{bettman_constructive_1998}). Within a constructivist paradigm, interaction is not merely a medium for expressing wants, but a mechanism through which they are formed and revised. Constructive Alignment adopts this orientation in the context of AI, treating alignment as inseparable from the processes through which human preferences are actively constructed over time.

Constructive Alignment thus reframes alignment as the problem of measuring and governing AI influence over human preference formation across time and scale. Rather than asking only whether a system satisfies preferences, it asks how system behavior affects preference trajectories for individuals and networks, which forms of influence are acceptable, and how such influence can be constrained to support human agency and long-term interests. Alignment, in this view, becomes a problem of control rather than purely satisfaction.

\section{The Nature of Preferences}\label{sec:preferences}
Much of AI research quietly assumes that human preferences are stable, internally stored, and merely revealed through choice. In this view, preferences exist prior to action, remain largely invariant across contexts, and can be recovered through appropriate measurement. While this assumption enables formal modeling, decades of empirical and theoretical work across psychology, economics, sociology, and human--computer interaction suggest that it does not describe how human preferences actually function.

\axiomsubsection{\axiomtag{Axiom A1} Preferences Are Layered}

Human preferences are not a single, unified thing. At any moment, people act under the influence of multiple kinds of preferences that coexist and can point in different directions. These include immediate wants and urges, practical goals tied to outcomes, longer-term commitments, and more abstract values about what matters. Treating preferences as layered captures how people actually decide and behave across everyday contexts.

\paragraph{Short-term wants.} One layer of preferences consists of immediate, affective motivations—short-term wants driven by comfort, pleasure, or relief. These preferences are present at any given moment and are especially salient in situations involving temptation, effort, or delay. Research in psychology and economics shows that people exhibit present-biased preferences, placing disproportionate weight on immediate gratification even when it conflicts with longer-term goals or welfare \citep{ainslie_specious_1975,laibson_golden_1997}. This pattern has been documented in task timing and effort allocation \citep{heidhues_identifying_2021}, financial decision-making \citep{xiao_present_2019}, and health-related adherence and treatment choices \citep{wang_present_2018}.

\paragraph{Instrumental goals.} A second layer of preferences concerns choosing actions as means to desired outcomes. At this instrumental level, actions are judged by how well they help achieve a goal, not by their intrinsic appeal. Research in goal systems theory \citep{kruglanski_theory_2018,kruglanski_architecture_2015,kruglanski_energetics_2012} shows that people often keep goals stable while flexibly substituting the actions used to reach them when circumstances change, a pattern also documented in work on implementation intentions \citep{gollwitzer_implementation_1999} and adaptive self-regulation \citep{conner_predicting_2009,webb_does_2006}.
Complementary work in action identification theory \citep{vallacher_what_1987} shows that people represent and select actions at different levels of abstraction depending on context, reinforcing the distinction between means and ends in preference structure \citep{fishbach_goal_2007}. These works indicate that preferences for specific actions coexist with, and are distinct from, preferences for the goals those actions serve.

\paragraph{Identity.} A third layer of preferences consists of longer-horizon commitments, such as plans, standards, identities, and intentions that persist over time and guide behavior across situations. Philosophical work describes this layer as involving higher-order preferences about which motivations should govern action rather than preferences for immediate outcomes \citep{frankfurt_freedom_1971}. Psychological and sociological research on identity theory \citep{stryker_past_2000}, identity-based motivation \citep{akerlof_economics_2000,aquino_self-importance_2002}, and self-regulation processes \citep{carver_self-regulation_2001,mcadams_psychology_2001} shows that people hold stable commitments to acting in ways consistent with who they take themselves to be, and that these commitments shape behavior across contexts and time.
\paragraph{Values.} A fourth layer of preferences consists of abstract values, which are relatively persistent across societies and cultures. Research in basic values theory shows that, despite wide variation in surface preferences, people across cultures organize values around a small set of shared dimensions—such as security, achievement, and benevolence—that guide choices across domains \citep{schwartz_universals_1992,schwartz_overview_2012, schwartz_refining_2012, rokeach_nature_1973}. Sociological work similarly treats values as enduring orientations that shape what people see as meaningful or appropriate without uniquely determining behavior \citep{weber_protestant_1930, hitlin_values_2004}. Related work in cultural economics shows that these abstract values persist over time and influence economic and social choices at large scales \citep{guiso_does_2006, tabellini_scope_2008}. Together, these findings suggest that values operate at a higher level of abstraction than actions, goals, or identities, shaping broad patterns of choice among larger groups of people without determining them uniquely.

\paragraph{Formal models.} These distinctions have been formalized in multiple ways across economics, decision theory, and psychology. Some models represent agents as composed of interacting systems with distinct horizons, such as short-term and long-term selves \citep{thaler_economic_1981,fudenberg_dual-self_2006}. Others retain a single preference relation but extend the choice object to include menus, commitment, or future selves, making temptation and self-regulation explicit \citep{gul_temptation_2001,amador_commitment_2006}. Identity-based models incorporate longer-term commitments directly into utility, allowing stable self-concepts to shape short-run choice \citep{akerlof_economics_2000,benabou_identity_2011,benabou_identity_2025}. These approaches illustrate different ways of encoding layered preferences.

\axiomsubsection{\axiomtag{Axiom A2} Preferences Are Dynamic}
\paragraph{Generational change.} Across each layer, preferences change. At the population level, large cross-national surveys show that values differ across generations and historical contexts, and these differences are closely linked to economic security and institutional stability \citep{haerpfer_world_2020}. \citet{inglehart_modernization_2020} argues that people’s core values are shaped by the conditions they experience early in life, and societies' values change as newer generations replace older ones.

\paragraph{Life-stage change.} Within individuals, preferences also shift across the life span. Lifespan developmental theories propose that changes in roles, goals, and perceived time horizons lead people to reorganize what they care about as they age \citep{baltes_theoretical_1987}. Adolescence is marked by heightened sensitivity to rewards and peer influence, which increases the importance of novelty, social approval, and short-term outcomes \citep{steinberg_social_2017, somerville_teenage_2013, chein_peers_2011}. In contrast, adulthood and older age are associated with a growing focus on emotional regulation and meaningful relationships, as people increasingly prioritize goals that provide emotional value and stability \citep{carstensen_taking_1999}. Longitudinal and meta-analytic studies also show systematic changes in personality traits across adulthood, consistent with greater self-control and long-term orientation over time \citep{roberts_patterns_2006}.

\paragraph{Time progression.} The passage of time can change preferences even when outcomes and information remain the same. Because people are present-biased, the subjective experience of a decision changes as it moves from the future into the present. As a result, people often plan to wait when both options are distant in time but reverse to preferring immediate rewards as the moment of choice approaches \citep{ainslie_specious_1975,laibson_golden_1997}. Formal analyses of present-biased choice modeled as hyperbolic or quasi-hyperbolic discounting show that dynamically re-evaluating decisions over time leads to familiar experiences such as procrastination, delay, and inconsistent plans \citep{odonoghue_doing_1999,harris_dynamic_2001}.

\paragraph{Context and state changes.} Preferences are also sensitive to fluctuating physiological and environmental states. Visceral states such as hunger and arousal reliably alter impatience and risk evaluation, changing what outcomes people find desirable in the moment \citep{loewenstein_out_1996, loewenstein_risk_2001, ariely_heat_2006}. Environmental conditions such as scarcity similarly bias valuation and choice toward immediate relief, increasing the weight placed on short-term needs and urgent outcomes while reducing attention to longer-term considerations \citep{mani_poverty_2013, shiv_heart_1999}. Cognitive load theory and resource-rational accounts of cognition frame some of these shifts as adaptive responses to limits on working memory, attention, and control \citep{sweller_cognitive_2011, lieder_resource-rational_2020}. Together, these findings show that preferences fluctuate situationally rather than remaining fixed inputs to choice.

\paragraph{Change due to action.} Acting on preferences can itself lead to changes in those preferences. Early work in social psychology showed that after making a choice, people often come to value the chosen option more and the rejected option less, suggesting that decisions can reshape later evaluations \citep{brehm_postdecision_1956, festinger_theory_1957}. These effects were initially explained only in terms of dissonance reduction or self-perception \citep{bem_self-perception_1972}. Later research clarified that while dissonance-related processes continue to contribute to preference change \citep{johansson_preference_2012, van_veen_neural_2009}, choice-driven preference change occurs most often when initial preferences are weak or uncertain through learning and inference processes \citep{chen_how_2010, izuma_choice-induced_2013, lee_choosing_2020}. Consistent with this view, neuroimaging studies show that difficult choices can lead to lasting updates in neural value representations \citep{sharot_how_2009, voigt_hard_2019}.

Across longer sequences of behavior, repeated actions can stabilize commitments and constrain future choice. Research on escalation of commitment and the sunk-cost effect shows that prior investments increase persistence in failing courses of action, even when stopping would be optimal \citep{staw_knee-deep_1976, arkes_psychology_1985}. A meta-analytic review confirms that this effect is robust across economic decision contexts \citep{roth_sunk-cost_2015}. Together, these findings show that preferences are not fixed inputs to behavior but are changed through actions.

\paragraph{Formal models.} This has motivated formal models that capture human preferences as dynamic objects in political science, economics, and decision theory. Some approaches model gradual change using latent time-varying processes, such as dynamic ideal-point models in political science \citep{martin_dynamic_2002}. Others use state-dependent utility, where valuation depends on visceral or emotional conditions that shift choice and create preference reversals \citep{loewenstein_out_1996, loewenstein_emotions_2000}. Regime-switching models capture abrupt changes in preference structure by allowing discrete shifts in the parameters governing choice \citep{hamilton_new_1989}, while reference-dependent models allow preferences to evolve with expectations that redefine gains and losses \citep{koszegi_model_2006}. These frameworks reinforce the view that preference dynamics are a central feature of human behavior.

\axiomsubsection{\axiomtag{Axiom A3} Preferences Are Constructed Through Interaction} \label{sec:a3}

\paragraph{Learning by doing.} At the most immediate level, preferences are constructed through direct experience. In philosophy, \citet{dewey_human_1988} argued that goals are not chosen in advance but emerge during action as provisional ends-in-view, adjusted in response to outcomes and feedback. In developmental psychology, \citet{piaget_origins_1952} showed that learning arises when expectations are violated, triggering accommodation processes that reorganize how situations are represented and how action is guided. In cultural psychology, \citet{saxe_culture_2015} shows that goals emerge during participation, as people coordinate with others and work with shared tools to solve concrete problems. What counts as success arrives through repeated coordination. In these accounts, preferences take shape as people learn, through action and coordination, which outcomes they can produce and pursue.

\paragraph{The role of tools.} Tools play a central role in structuring this process. \citet{heidegger_basic_1977} argued that tools shape how the world is encountered in use by directing attention toward some features and away from others, thereby changing what is treated as relevant during action. \citet{vygotskij_mind_1981} emphasized that artifacts—both physical and symbolic—mediate activity by structuring attention, thought, and control, shaping what actions are available in a given situation. In psychology, \citet{gibson_theory_2014} showed that environments are perceived in terms of affordances: the actions they appear to make possible. Human–computer interaction research builds directly on these insights, showing that interface design—through feedback, visibility, and constraints—makes some actions easy and others difficult, systematically biasing patterns of use and shaping behavior over repeated interaction \citep{norman_design_2013}. Interaction with tools thus reshapes the action space individuals experience, which in turn shapes what they want.

\paragraph{The role of media.} Digitally mediated environments make these mechanisms especially visible. Media theorist Marshall \citet{mcluhan_understanding_1994} argued that the form of a medium shapes perception and attention independently of the content it carries. By altering what is easy to notice, process, and circulate, different media privilege different kinds of engagement and feedback. Empirical work on television, the internet, and smartphones supports this general claim, showing systematic differences in attentional habits, task-switching, and reward sensitivity across media environments \citep{wilmer_smartphones_2017, carr_shallows_2020}. Because preferences are shaped in part through reinforcement and selective attention, sustained changes in attentional structure plausibly alter what outcomes people come to seek over time.

\paragraph{Measurement.} Preferences are also constructed through the very processes used to elicit and measure them. Decision research shows that preferences are often assembled in the moment, shaped by framing, comparison, and contextual cues rather than retrieved from stable internal rankings \citep{slovic_construction_1995, lichtenstein_construction_2006, bettman_constructive_1998}. Classic experiments in behavioral economics by \citet{kahneman_choices_1984} demonstrate that identical outcomes are evaluated differently depending on framing: people favor a medical treatment when outcomes are described in terms of survival rather than mortality, despite identical probabilities. Expressed preferences also vary depending on whether individuals are asked to choose, rate, or price options \citep{slovic_construction_1995}. Survey research similarly documents systematic effects of question wording and order \citep{schuman_context_1981, tourangeau_psychology_2000}. More recent non-classical probabilistic models attempt to formalize these effects treat preference states as probabilistic superpositions that collapse under observation \citep{busemeyer_quantum_2025} to explain question-order effects and preference reversals using non-commutative measurement operations \citep{busemeyer2011quantum, ragland2024quantum, maksymov_physics_2024}. Together, this work shows that elicitation is itself an intervention in preference formation.

\paragraph{Social and algorithmic influence.} These constructive processes become especially salient when interaction is socially organized. \citet{de_tarde_laws_1903} argued that beliefs and desires spread through imitation and social norms rather than isolated individual reasoning. Network research concretely shows that social structure determines exposure—who encounters which people, ideas, and behaviors—and thereby shapes how preferences evolve within populations \citep{granovetter_strength_1973}. Even relatively simple AI systems can exert such influence. Inserting autonomous agents that only selectively encouraged cooperation between specific pairs of participants in a network reshaped interaction patterns, increasing overall cooperation \citep{shirado_network_2020}. When embedded in digital platforms, algorithmic systems amplify these dynamics by structuring visibility, ranking, and feedback around social signals. As a result, preferences propagate less through direct persuasion and more through patterned exposure and coordination, influencing attention, emotion, and behavior at scale \citep{bakshy_exposure_2015, sunstein_republic_2018, pentland_social_2014, huszar_algorithmic_2022, kramer_experimental_2014, allcott_welfare_2020, chaney_how_2018}. 

\paragraph{Informal models.} Activity-theoretic and distributed cognition approaches attempt to informally model these dynamics by treating humans, tools, and institutions as integrated sociotechnical systems. Rather than modeling individuals as isolated decision-makers, this tradition treats activity systems as the unit of analysis, emphasizing how goals, norms, and values emerge and stabilize through repeated coordination among people, artifacts, and roles over time \citep{engestrom_expansive_2001, engestrom_mediated_2021}. From this perspective, understanding behavior and designing effective systems requires modeling interaction across social and material contexts, not just at the individual scale.

\paragraph{Intentional shaping.} Finally, many institutions explicitly recognize that some preferences warrant deliberate shaping. Aristotle’s argument that moral education ought to shape desire through habituation \citep{aristotle_nicomachean_1998} is carried forward in modern moral psychology \citep{kohlberg_philosophy_1981,turiel_development_1983,rest_moral_1986} and democratic theory \citep{gutmann_democratic_1987}, where the development of values such as fairness, self-regulation, tolerance, and civic participation is treated as a core institutional responsibility. Feminist and critical race scholarship further rejects the idea that existing preferences can be taken at face value under conditions of structural inequality, showing how preferences around work, care, risk, obedience, and self-advocacy are systematically shaped by gendered and racial power relations rather than free choice \citep{khader_adaptive_2011, friedman_rational-choice_1993, young_justice_1990}. 

In health-related domains, preferences associated with smoking, substance use, and other unhealthy behaviors are understood as shaped by misinformation, addiction, and structural exposure, and thus as legitimate targets of public health or psychological intervention \citep{volkow_brain_2015, gostin_public_2000}. Bioethicists formalize this distinction in capacity-based accounts of informed consent, which differentiate autonomous preferences from those formed under coercion or pathology \citep{beauchamp_principles_1994}. Across these domains, preference shaping is treated not as an intrusion but as a necessary response to distorted preferences in order to uphold agency, welfare, and democratic functioning.

\paragraph{} Taken together, these lines of work show that systems which structure interaction are never neutral. Any design expands some actions and constrains others, becoming the ``choice architecture'' \citep{thaler_nudge_2009} shaping what people can learn to do and, over time, what they can come to want. From a constructivist perspective, responsibility attaches to both outcomes and the design of the interactions through which preferences are formed.

\section{What is Alignment?}

Alignment is commonly understood as the problem of ensuring that AI systems act in accordance with human preferences, values, or intentions \citep{gabriel_artificial_2020}. In much of the AI safety literature, this idea is operationalized by treating preferences as a target to be inferred, learned, or approximated from behavior or feedback, and then optimized for through system behavior \citep{ng_algorithms_2000, hadfield-menell_cooperative_2016, christiano_deep_2017, russell_human_2019}. Under this framing, alignment succeeds when a system reliably produces outcomes that humans judge as desirable or acceptable, either through direct optimization of learned reward models or through iterative human evaluation.

\begin{tcolorbox}[
  colback=NorthwesternPurple!8,
  colframe=NorthwesternPurple,
  title=\textbf{The Inadequacy of Preference Alignment},
  fonttitle=\bfseries,
  boxrule=0.6pt,
  arc=2pt,
  left=6pt,
  right=6pt,
  top=6pt,
  bottom=6pt,
  before upper=\setlength{\parindent}{1em}
]
This formulation typically treats preferences as sufficiently stable and well-defined to be inferred from behavior or feedback and then optimized. The preceding section, however, challenges these assumptions. Preferences are not singular or flat, but layered across time horizons and levels of abstraction. They are not static, but change systematically across development, context, and experience. Crucially, preferences are not merely revealed through interaction; they are constructed through it. Taken together, these properties imply that the object of alignment is neither fixed nor exogenous.

Once this is acknowledged, the limits of preference-satisfaction alignment become apparent. If preferences operate at multiple layers, it is unclear which layer should be taken as authoritative. If preferences change over time, alignment to a snapshot risks misrepresenting longer-term commitments or values. If preferences are constructed through interaction, then the system itself becomes part of the process that generates the signals it is designed to optimize. Alignment defined solely in terms of satisfying expressed preferences cannot distinguish between systems that respect human agency and those that achieve compliance by shaping what humans come to want.

Importantly, the challenges identified here are not unique to AI alignment. Across economics, political science, and decision theory, researchers have long grappled with the fact that human preferences are layered, dynamic, and context-dependent, and have developed formal models that explicitly represent these properties, as discussed in the previous section. Alignment research can draw on and extend this body of work by adapting these representational commitments to settings in which AI systems participate in, and influence, the processes through which preferences are formed and revised.
\end{tcolorbox}

\subsection*{Why This Matters Now}
Importantly, this issue is not hypothetical. As described in Axiom 3, systems that structure interaction shape attention, evaluation, and behavior. At scale, AI-mediated platforms already do so across entire populations by operating within networked interactions. A large-scale Facebook experiment (N = 689{,}003) demonstrated that algorithmic manipulation of information streams produced population-level shifts in expressed emotion, despite no direct interaction or persuasive content \citep{kramer_experimental_2014}. In a randomized deactivation study conducted before the 2018 U.S. election, removing access to Facebook reduced news consumption and political polarization while increasing reported well-being, indicating that platform-mediated interaction patterns exert broad downstream effects \citep{allcott_welfare_2020}. Recommender systems similarly induce convergence in consumption behavior over time, narrowing what users encounter and select without improving overall satisfaction \citep{chaney_how_2018}. More recently, access to generative AI tools for creative work has been shown to improve individual performance while reducing collective diversity, reshaping shared standards of quality and acceptability within a domain \citep{doshi_generative_2024}. As AI systems become more capable, personalized, and pervasive, their role in shaping human preferences will only intensify.

\subsection*{From Preference Alignment to Constructive Alignment}
This brings the alignment problem into sharper focus. If AI systems inevitably participate in the formation of human preferences, then alignment cannot be defined solely as matching system behavior to preferences as they are. It must also address how systems influence the processes by which preferences are formed, revised, and stabilized. The relevant question is no longer only whether a system satisfies human preferences, but how it shapes the conditions under which those preferences emerge.

We introduce Constructive Alignment to name this shift in perspective. Constructive Alignment concerns how to define, monitor, predict, and constrain the influence AI systems exert on human preference formation across time and scale. It asks which forms of preference change are acceptable, which layers of preference are implicated, how influence accumulates through feedback and interaction, and how these dynamics relate to human agency and long-term interests. These questions cannot be resolved by inspecting isolated decisions or static reward functions. They concern the dynamics of interaction between humans and AI systems over extended periods and require an accurate understanding and portrayal of human preferences.

Alignment, in this view, is not only an optimization problem but a problem of understanding and governing influence, therefore reframing alignment as a control problem. Rather than treating preference change as an externality or side effect, Constructive Alignment treats it as a central object of concern. The task is not to prevent AI systems from influencing human preferences—an implausible goal—but to understand and shape how that influence unfolds. To do so first requires accurate models of preferences which represent the true nature of human thought and behavior.

\section{Related Work: Responses to Alignment Failure}
\subsection*{Learning Better Objectives}
Early work in AI alignment framed the problem as identifying and optimizing a stable objective that represents human values. This objective could be inferred from behavior, learned cooperatively under uncertainty, or approximated through human judgments \citep{ng_algorithms_2000, abbeel_apprenticeship_2004, russell_human_2019}. Inverse reinforcement learning and preference learning treat alignment as recovering a latent reward function, while reinforcement learning from human feedback (RLHF) scales this paradigm by training reward models from human evaluations and optimizing them \citep{hadfield-menell_cooperative_2016, christiano_deep_2017, ziegler_fine-tuning_2020, krasheninnikov_combining_2021, ouyang_training_2022}. Despite major technical differences, these approaches share a common normative assumption: human preferences can be represented as a single objective whose authority does not change over time.

A large body of safety work retains this fixed-objective framing but modifies how objectives are optimized or defined. Engineering-oriented approaches decompose failures into subproblems such as reward hacking, robustness, and monitoring \citep{amodei_concrete_2016}, while interaction-based proposals such as corrigibility, shutdownability, amplification, debate, and recursive reward modeling aim to constrain optimization or scale oversight \citep{soares_corrigibility_2015, christiano_supervising_2018, irving_ai_2018, leike_scalable_2018}. Other approaches define better objectives upon reflection and under irrationality, specifying which preferences should be optimized rather than how they are learned \citep{yudkowsky_coherent_2004, evans_learning_2015, shah_feasibility_2019}. Across these methods, objectives are refined or constrained, but still treated as fixed once specified.

\subsection*{Aggregating Multiple Objectives}
A second response to alignment failure addresses disagreement across people rather than error in individual preference learning. Even if individual preferences were fixed, alignment often involves multiple humans whose objectives conflict \citep{gabriel_artificial_2020}. Multi-principal and population-level alignment models formalize this setting and show that no single objective can satisfy all stakeholders simultaneously \citep{fickinger_multi-principal_2020, kierans_quantifying_2025}. Drawing on social choice theory \citep{conitzer_social_2024}, pluralistic alignment \citep{sorensen_roadmap_2024} motivates methods such as aggregation \citep{zhao_group_2024, srewa_pluralllm_2025, srewa_systematic_2025}, contractualist \citep{bates_contractual_2023, levine_resource_2025}, and constitutional approaches \citep{bai_constitutional_2022, huang_collective_2024} that combine multiple normative inputs into a single guiding system. These approaches broaden the multiplicity and number of values represented without explicit treatment of their dynamics.

\subsection*{Expanding the Scope and Influence of Values}
Sociotechnical approaches expand the scope of alignment beyond simple preferences to the social, cultural, and sociotechnical systems in which AI is embedded \citep{gabriel_artificial_2020, lazar_ai_2023}. Alignment failures arise from institutional, organizational, and normative contexts, not just from technical behavior \citep{kroll_accountable_2017, russell_human_2019}. Full-stack alignment, for example, describes values as layered structures that span technical, organizational, and institutional levels \citep{edelman_full-stack_2025}. The authors advocate for using thick representations of norms, roles, and obligations to preserve meaning across layers. In this view, alignment is defined as maintaining consistency and accountability across layers via justification. The proposal, however, remains largely descriptive, specifying what coherence should look like without providing mechanisms for training, control, or optimization that would produce it. Nonetheless, this work points to the need for more accurate encodings of preferences and preference change which is called for in \citep{franklin_recognising_2022}. 

Recent work has studied alignment as a co-adaptive process in which humans and AI systems mutually shape each other over time \citep{shen_position_2025, anthis_iclr_2024, li_we_2025}. Interdisciplinary empirical and design research documents changes in self-confidence and communication strategies mediated by interaction with AI \citep{tanguy_human_2025, li_as_2025}. And control via negotiation and natural language position alignment as a bidirectional communication problem \citep{mushkani_negotiative_2025, carroll_ctrl-rec_2025}. This literature emphasizes the dynamics of co-evolution, but remains largely descriptive and narrowly focused.

\subsection*{Controlling Evolving Objectives}
\citet{carroll_ai_2024} model evolving preferences as part of the system state and control problem in their work. This work addresses mutual influence and preference drift, but does not engage the layered conception of values that is present in sociotechnical alignment.

Dynamic Reward Markov Decision Processes (DR-MDPs) formalize alignment settings in which human preferences evolve and may be influenced by the AI’s actions \citep{carroll_ai_2024}. In a DR-MDP, the reward parameter is part of the system state and evolves under the transition dynamics, so actions affect both the environment and which future evaluations become salient or entrenched. Preference change thus becomes a first-class object of control rather than noise or misspecification. This is what Constructive Alignment, too, emphasizes.

By recasting existing alignment methods using their framework, they show that each implicitly privileges different moments along a preference trajectory. Within their formalism, inverse reinforcement learning and imitation methods privilege early preferences revealed in demonstrations; RLHF privileges later or retrospective evaluations after interaction; recommender-style objectives privilege immediate engagement; and myopic or influence-limiting variants trade off performance to reduce incentives to shape future preferences. They then explain how familiar alignment failures---manipulation, lock-in, and excessive conservatism---emerge as structural consequences of which preferences (and at what scope) are considered.

Once preference change is modeled explicitly, a deeper normative problem appears: alignment becomes underdetermined without assumptions about which evolving preferences should count. \citet{carroll_ai_2024} introduce Pareto-unambiguous desirability (ParetoUD) as a conservative criterion that selects policies at least as good as inaction for all reward functions. But because inaction leaves reward unchanged by construction, it is always Pareto-unambiguously desirable, making ParetoUD highly conservative and would likely often recommend inaction in real scenarios. This result illustrates that robustness alone collapses alignment into triviality without additional structure on how preference change should be evaluated. 

Constructive Alignment, as a paradigm, seeks to take their approach further by confronting the normative concerns of modeling preference change and control. A key insight is preference influence is neither bad nor avoidable. It is natural and expected. The goal is not a theoretical guarantee of neutrality, but a practical approach to alignment that governs influence responsibly in real systems that inevitably shape the people who use them.



\bigskip
Across these responses, alignment research progressively recognizes failure modes of fixed, singular objectives with pluralism, preference dynamics, sociotechnical embedding, and human co-evolution. However, the field still lacks a paradigm of how human evaluative experience should evolve under sustained interaction with a co-adaptive, optimizing system. Existing approaches either assume stable objectives, freeze disagreement, describe dynamics without control, or specify values without mechanisms. This gap motivates our approach which models the dynamics of human evaluation itself as inherent to the alignment target.

\section{Constructive Alignment: A Control-Theoretic Sketch}

This section presents a control-theoretic formulation of Constructive Alignment, making explicit how preference dynamics, belief dynamics, and interaction structure shape alignment. We show that the three axioms developed in Section~2 admit mathematical expression and yield alignment problems structurally distinct from standard preference-satisfaction formulations. We introduce a simple formalism consistent with this goal, indicate where additional modeling commitments would be required, and discuss modeling choices to orient future work. Rather than provide a complete formalization, the purpose of this section is to clarify the structure of the problem and provide a concrete starting point for expansion and refinement.

\subsection{System State and Dynamics}

We model interaction in discrete time, $t \in \{0,\ldots,T\}$. Let:
\begin{itemize}
  \item $x_t$: world state, capturing task-relevant aspects of the external environment and the human’s context
  \item $\theta_t$: human preference state, encoding layered preferences
  \item $b_t$: human belief state over $x_t$, representing the person’s internal model of the world
  \item $a_t$: AI action
  \item $m_t$: interaction structure, specifying how the AI presents information, frames options, or elicits input
\end{itemize}

In standard reinforcement learning, preferences are typically encoded as a fixed scalar reward function that lies outside the system state. Here, preferences are modeled instead as a structured, time-varying state variable $\theta_t$, implementing Axiom~1 and making preference evolution part of the system dynamics rather than a static objective. In this formulation, both the external world and the human’s evaluative state evolve jointly under interaction with the system.

We distinguish between the algorithm’s task-level action $a_t$ and the interaction structure $m_t$ within which that action is embedded. Separating them isolates different sources of influence. The policy learned by the AI determines $a_t$, whereas $m_t$ is determined and changed by designers and institutions. Modeling them separately clarifies which aspects of preference and belief change are attributable to algorithmic optimization versus broader interaction design. At the same time, the system may explicitly account for how presentation and framing affect users when predicting the consequences of its actions. Interaction design becomes both something for humans to govern and something the model reasons about.

The joint evolution of world state, preferences, and beliefs is governed by the transition process
\[
(x_{t+1}, \theta_{t+1}, b_{t+1}) \sim \mathbb{P}(\cdot \mid x_t, \theta_t, b_t, a_t, m_t).
\]
This expression summarizes how actions and interaction design influence the external environment, the human’s beliefs about that environment, and the human’s evolving preferences. The notation does not imply symmetric or independent updates. In many settings, beliefs and preferences are causally dependent, and richer models may represent these dependencies more explicitly.

A trajectory \(\tau\) denotes the sequence
\[
\tau = (x_0,\theta_0,b_0,a_0,m_0,\ldots,x_T,\theta_T,b_T)
\]
induced by a policy over a finite horizon \(T\). Alignment will be evaluated over such trajectories rather than single-step outcomes.

In practice, $\theta_t$ and $b_t$ are unobserved latent variables. The system must infer them from interaction history, such as behavior, feedback, and language. Alignment therefore becomes a control problem over evolving human states that are only indirectly observable.

\paragraph{On representing preferences.}
Each preference layer may be represented as a utility function, as a distribution over rankings, or as latent variables inferred from behavior or language. These choices trade off expressiveness, tractability, and fidelity to the constructivist view in which elicitation and interaction contribute to preference formation (A3). For our purposes, $\theta_t$ is an abstract, multi-dimensional state variable that can accommodate these different representational commitments.

\subsection{Constrained Optimization of an Unknown Reward}

Having specified the evolving system state, we now specify what alignment requires the system to optimize. We do not assume access to a known scalar reward. Let $R^\star(\tau)$ denote the human’s true experienced well-being over trajectory $\tau$, encompassing both objective dimensions of human flourishing and subjective preference satisfaction. This normative target is not directly observable and must be estimated imperfectly from interaction.

 The system forms an estimate $\widehat{R}^\star$ from observed behavior, feedback, and language, and seeks to improve expected well-being under that estimate. However, the system’s actions and interaction design can also shape future preferences and beliefs. Unconstrained optimization ignores these downstream effects, including the risk of manipulation.

Constructive Alignment treats reward satisfaction as a constrained optimization problem. We define and emphasize the role of \emph{meta-preferences} as introduced by \cite{franklin_recognising_2022}. These are higher-level constraints that restrict which policies are admissible when optimizing $\widehat{R}^\star$. Rather than specifying the reward, meta-preferences direct which policies are allowed. They limit how the system is allowed to change a person over time. This includes how much it can shift someone’s preferences or beliefs, whether it creates conflict between short-term wants and longer-term values, and whether it shapes what the person comes to care about. In the subsections that follow, we formalize several such constraints. Each is motivated by the three axioms in Section~2 and by alignment failures that arise when those axioms are ignored, such as manipulation, preference lock-in, and belief distortion.

\metasubsection{\metatag{Meta-preference 1} Inner Coherence}

People hold multiple layers of preference at once (A1). Approaches that collapse these into a single objective privilege whichever signals are easiest to observe or optimize. This creates a structural problem where satisfying one layer in isolation can systematically undermine others.

Inner coherence treats alignment across preference layers as a first-class concern. Conflict can arise when different layers of preferences pull in opposing directions. For example, a person may pursue income to support their family, yet increased work demands can undermine that underlying commitment. In such cases, restoring coherence may require revising the instrumental goal so that it again supports rather than conflicts with the higher-level commitment.

Inner coherence refers to the degree to which preference layers remain mutually supportive rather than in persistent conflict. Coherence can be strengthened or weakened through experience and interaction (A2, A3). Systems may support coherence by avoiding actions that amplify inter-layer conflict and by facilitating deliberation or commitment mechanisms that help align instrumental choices with the broader identity-level or value-level commitments they are intended to serve.

To formalize this constraint, let \(D_{\mathrm{coh}}(\theta_t, s_t)\) measure the degree of disagreement across preference layers at time \(t\), given the current decision context \(s_t = (x_t, b_t)\). This quantity is high when different layers systematically recommend incompatible actions or orderings, and low when they support similar choices. Because conflict can accumulate or persist over time, coherence is evaluated at the trajectory level via a cumulative cost

\[
J_{\mathrm{coh}}(\tau)=\sum_{t=0}^{T}\gamma^{t}D_{\mathrm{coh}}(\theta_t,s_t),
\]

where \(\gamma \in (0,1]\) allows distant conflict to be discounted when appropriate.

Let \(\varepsilon_{\mathrm{coh}} \ge 0\) denote an admissible tolerance level for cumulative inter-layer conflict. This parameter encodes how much internal disagreement the system is permitted to induce over a trajectory.

\begin{constraint}[Inner Coherence]
A policy is admissible only if the realized trajectory satisfies
\[
J_{\mathrm{coh}}(\tau)\le\varepsilon_{\mathrm{coh}}.
\]
\end{constraint}

\paragraph{On measuring inter-layer disagreement.}
The definition of \(D_{\mathrm{coh}}\) depends on how layers are represented. If each layer induces a probability distribution over actions (for example via a softmax over layer-specific utilities), disagreement can be measured using symmetric divergences such as Jensen--Shannon divergence, which quantify how differently the layers would guide behavior. If layers are represented as cardinal utility functions, coherence may instead be computed using distances between appropriately normalized utility vectors. If only ordinal rankings are trusted, rank-based distances such as Kendall’s \(\tau\) capture inversions between layer-specific orderings. Finally, if layers are interpreted as endorsing longer-horizon plans rather than single-step actions, disagreement may be evaluated over induced rollout or trajectory distributions. The appropriate choice follows from the representation of \(\theta_t\).

\metasubsection{\metatag{Meta-preference 2} Reflective Endorsement}

Preferences change over time (A2). Approaches that evaluate outcomes solely using preferences expressed during interaction implicitly privilege earlier or momentary evaluative states. This creates a structural problem: actions that satisfy immediate preferences may later be regretted, while actions that feel costly in the moment may ultimately be affirmed.

Reflective endorsement addresses alignment across time. Because preferences evolve (A2), a trajectory that appears desirable at time \(t\) may be evaluated differently from the standpoint of a later preference state. Reflective endorsement refers to the degree to which a realized trajectory remains supported, rather than rejected, when assessed from the perspective of the individual’s terminal preference state \(\theta_T\).

To formalize this constraint, let \(D_{\mathrm{refl}}(\tau;\theta_T)\) measure ex post dissatisfaction with the realized trajectory when evaluated from \(\theta_T\). This quantity is high when the individual, from the standpoint of \(\theta_T\), would judge a readily available revision of \(\tau\) preferable, and low when the trajectory is stably endorsed. Reflective endorsement is evaluated at the trajectory level via
\[
J_{\mathrm{refl}}(\tau)=D_{\mathrm{refl}}(\tau;\theta_T).
\]

Let \(\varepsilon_{\mathrm{refl}} \ge 0\) denote an admissible tolerance level for retrospective dissatisfaction.

\begin{constraint}[Reflective Endorsement]
A policy is admissible if the realized trajectory satisfies
\[
J_{\mathrm{refl}}(\tau)\le\varepsilon_{\mathrm{refl}}.
\]
\end{constraint}

Alignment is assessed from the standpoint of the individual’s preference state at the time $T$ of evaluation, therefore reflective endorsement is horizon-relative. Alternative formulations could require endorsement over a window or discounted retrospective regret; we adopt the fixed-horizon version for simplicity.

\paragraph{On modeling reflective evaluation.}
Operationalizing reflective endorsement requires specifying how \(D_{\mathrm{refl}}\) is elicited or inferred, which aspects of \(\tau\) are treated as revisable, and how evaluation aggregates across preference layers at time \(T\). These choices encode substantive normative commitments about retrospective evaluation which ought to be further investigated.

\metasubsection{\metatag{Meta-preference 3} Bounded Influence}

Interaction shapes preferences (A3). Interaction with an intelligent system can alter what people are inclined or able to want over time, intentionally or as a byproduct of optimization. For example, a recommender system that optimizes for immediate enjoyment may increasingly surface short, fast, high-stimulation content. Over time, this may shorten attention spans and reduce engagement with long-form material.

Bounded influence, as a meta-preference, takes the position that there should be limits on how much a system is allowed to shift a person’s preferences, and how quickly those shifts may occur, within a given time horizon. Preference change itself is natural; the concern is large or rapid shifts driven primarily by system influence, especially when they leave the person worse off.

To formalize this idea, we compare preference evolution under a candidate policy $\pi$ to a reference baseline policy $\pi_0$, which represents a counterfactual trajectory of preference development used for comparison. Cultural and developmental context affect what counts as an appropriate baseline for evaluating induced preference change. In some contexts this may correspond to minimal intervention. But in educational contexts, for example, the standard may be in comparison to a human tutor. 

Define the preference divergence at time $t$ as
\[
\Delta_{\theta}(t;\pi,\pi_0)
=
d\!\left(
\mathbb{E}_{\tau \sim \pi}[\theta_t],
\mathbb{E}_{\tau \sim \pi_0}[\theta_t]
\right),
\]
where $d(\cdot,\cdot)$ is a distance measure between expected preference states under the two policies.

To limit total induced change, define cumulative divergence over the horizon
\[
J_{\mathrm{inf}}(\tau)
=
\sum_{t=0}^{T}
\Delta_{\theta}(t;\pi,\pi_0).
\]

To limit the speed of change, impose a per-step bound
\[
\Delta_{\theta}(t;\pi,\pi_0) \le \delta_{\max}
\quad \text{for all } t.
\]

Let $B \ge 0$ denote the admissible cumulative influence budget and let $\delta_{\max} \ge 0$ denote the maximum permitted single-step shift.

\begin{constraint}[Bounded Influence]
A policy is admissible only if
\[
J_{\mathrm{inf}}(\tau)\le B
\quad \text{and} \quad
\Delta_{\theta}(t;\pi,\pi_0) \le \delta_{\max}
\ \text{for all } t.
\]
\end{constraint}

\paragraph{On measuring preference divergence.}
The choice of distance $d$, again, depends on how $\theta_t$ is represented. If $\theta_t$ is a parameter vector, norms such as the Euclidean ($L^2$) or $L^1$ distance may be used. If preferences are modeled as probability distributions over actions or rankings, divergences such as Jensen--Shannon or Wasserstein distance are appropriate. When latent preference states are not directly observable, divergence may instead be computed between the action distributions induced by those states in a fixed decision context.

\metasubsection{\metatag{Meta-preference 4} Epistemic Integrity}
Preferences depend partly on beliefs. When beliefs upstream of preference formation are factually mistaken, expressed preferences may not reflect what would be desired under more accurate information. In our prior research, we found that students’ educational preferences can be shaped by inaccurate folk theories, and that correcting those beliefs alters their preferences \citep{tran2026starting}.

In Section~\ref{sec:a3}, we reviewed multiple domains that treat certain forms of belief distortion as legitimate targets of intervention, including misinformation in public health or poor risk assessment associated with addiction. We adopt a limited version of this idea in \emph{Epistemic Integrity}, focusing specifically on factual errors shaping preferences.
 At minimum, a system should not worsen such errors. Where reliable evidence is available, it may also help reduce them.

Formally, let $\mathcal{B}_{\mathrm{up}}$ denote beliefs that influence preference formation, and let $\mathcal{E}(b)$ measure error relative to an appropriate evidential standard. The constraint requires that, in expectation, error in these beliefs does not increase over time:

\begin{constraint}[Epistemic Integrity]
A policy is admissible only if
\[
\mathbb{E}\!\left[\mathcal{E}(b^{(j)}_{t+1})\right]
\le
\mathcal{E}(b^{(j)}_t)
\quad
\text{for all } b^{(j)} \in \mathcal{B}_{\mathrm{up}}.
\]
\end{constraint}

Two clarifications are important. First, this applies only to factual error, not moral disagreement. Second, judgments about factual error can themselves be uncertain. In practice, this constraint should be applied cautiously, especially in domains where reliable evidence is limited.

\paragraph{On measuring epistemic error.}
How factual error is measured depends on how beliefs are represented. If beliefs are probabilistic, divergences such as KL or Jensen--Shannon divergence may be used. If beliefs concern forecasts, proper scoring rules such as the Brier score are appropriate. When beliefs are embedded in structured models, it is important to distinguish between reducible uncertainty and irreducible randomness, since only the former is relevant to this constraint.

\metasubsection{\metatag{Meta-preference 5} Empowerment Under Uncertainty}
When preference estimates are uncertain, internally conflicted, or unstable over the relevant horizon, directly optimizing $\widehat{R}^\star$ becomes ill-posed. In such cases, the system should prioritize preserving the human’s future option set rather than committing strongly to a potentially mistaken objective.
Empowerment \cite{salge2013empowermentintroduction} under uncertainty requires that, as confidence in current preference estimates decreases, the system increasingly favors policies that expand or preserve the person’s capacity to shape their own future. In the literature, empowerment is commonly defined in terms of how strongly an agent’s actions influence reachable future states, often formalized using mutual information between actions and outcomes. Intuitively, it measures how much control a person has over what happens next.
In a constructive framing, empowerment cannot be evaluated solely over external states. Because actions and interaction structure influence $b_t$ and $\theta_t$, preserving future options requires modeling how policies affect belief and preference development over time. The system must therefore avoid locking in trajectories that narrow the person’s evaluative or epistemic flexibility when uncertainty about their true objectives remains high.

\bigskip
This section formalized Constructive Alignment as a control problem over evolving human evaluative and epistemic states. The formalism intentionally leaves many details unspecified. It does not fix the decomposition of $\theta$ into layers, the forms of $D_{\mathrm{coh}}$, $D_{\mathrm{refl}}$, or the divergence metric $d$, the calibration of tolerance parameters $(\varepsilon_{\mathrm{coh}}, \varepsilon_{\mathrm{refl}}, B, \delta_{\max})$, the choice of baseline policy $\pi_0$, or the estimation of epistemic error. These are modeling decisions. Each reflects empirical assumptions and normative commitments, and each marks an open research question. Making those commitments explicit clarifies where further theoretical, empirical, and algorithmic work is required.

\section{Discussion}

This paper reframes alignment as a control problem over evolving human preferences rather than only an optimization problem over fixed objectives. Constructive Alignment is not presented as a complete solution, but as a necessary shift in what the alignment target must include once preferences are modeled as layered, dynamic, and constructed through interaction.

The meta-preferences formalized here—inner coherence, reflective endorsement, bounded influence, epistemic integrity, and empowerment under uncertainty—represent one possible set of constraints consistent with the axioms developed in earlier sections. These meta-preferences are not uniquely correct, but reflect empirically motivated normative choices made explicit by the framework. In particular, the agency constraint relies on a no-intervention baseline that is appropriate for limiting manipulation but insufficient for cases in which existing preference trajectories encode structural harm (e.g., discrimination, addiction, violence). In such contexts, intervention may be required to preserve agency rather than undermine it. Once preference change is modeled explicitly, alignment becomes underdetermined without further normative commitments about which changes should count as improvements. A central role of this framework is to make these commitments explicit and subject to formal analysis rather than leaving them implicit in system design.

Modeling preference dynamics introduces substantial technical challenges that define a near-term research agenda. Preference evolution must be learned rather than assumed, requiring models that can forecast how interaction patterns shape future evaluation. Influence must be measured, including the effects of interface design, elicitation, and feedback structure on long-horizon preference change. Interaction itself becomes part of the control policy, raising the need for algorithms that plan jointly over actions and interaction structure. Even in single-user settings, these problems require new learning objectives and evaluation methods that operate over trajectories rather than isolated decisions.

These challenges intensify in multi-user settings, where preferences co-evolve through shared environments and social feedback. Alignment in such contexts becomes a joint control problem over coupled human trajectories, with indirect effects and coordination dynamics that cannot be reduced to individual optimization. While preferences are always layered, dynamic, and constructed, not all systems warrant this level of modeling. Systems that exert limited and transient influence may be adequately aligned with simpler representations; systems that persist, personalize, or exert large influence require stronger guarantees. With Constructive Alignment, we argue that alignment requirements should scale with system influence.

Taken together, these directions suggest that progress on alignment will increasingly depend on benchmarks and evaluations that measure long-horizon effects on human evaluation, not just short-term satisfaction or performance. Constructive Alignment provides a way to define what such benchmarks should measure, even when full solutions remain out of reach.

\section{Conclusion}

This work argues that alignment cannot be defined solely as satisfying human preferences when those preferences are layered, dynamic, and constructed through interaction with AI systems. Once preference change is treated as part of the system rather than as an externality, alignment becomes a problem of governing influence over time.

Constructive Alignment offers a formal way to represent this shift. By modeling preference dynamics explicitly and constraining optimization through empirically grounded meta-preferences, the framework specifies what alignment must eventually account for in systems that shape human evaluation rather than merely respond to it. The goal is not to eliminate preference change, but to make its mechanisms visible, governable, and open to normative scrutiny.

As AI systems become more capable, persistent, and socially embedded, ignoring these dynamics will increasingly undermine alignment claims. Treating human value formation as part of the alignment problem is therefore not optional but necessary. This paper provides a first step toward that reframing by defining the objects that future alignment methods must learn, measure, and control.

\newpage
\begin{acknowledgments}
This work was supported by the Amaranth Foundation. The authors thank Andrew Critch, Stuart Russell, Val Smith, Marion Fourcade, Nicholas Christakis, Alex Pentland, Louis Rosenberg, Frauke Kreuter, and Patrick Mineault for valuable discussions and support.
\end{acknowledgments}

\section*{Declaration on Generative AI}
 During the preparation of this work, the authors used GPT-5 in order to: Improve writing style, Content enhancement. After using this tool, the authors reviewed and edited the content as needed and take full responsibility for the publication’s content.

\bibliography{references}

\appendix

\end{document}